\documentclass[letterpaper]{article} 
\usepackage{aaai2026}  
\usepackage{times}  
\usepackage{helvet}  
\usepackage{courier}  
\usepackage[hyphens]{url}  
\usepackage{graphicx} 
\usepackage{natbib}  
\usepackage{caption} 
\usepackage{algorithm}
\usepackage{algorithmic}

\usepackage{newfloat}
\usepackage{listings}
\DeclareCaptionStyle{ruled}{labelfont=normalfont,labelsep=colon,strut=off} 
\floatstyle{ruled}
\newfloat{listing}{tb}{lst}{}
\floatname{listing}{Listing}
\usepackage{xspace}

\usepackage{amsmath}

\usepackage{subfigure}
\usepackage{subfig}
\usepackage{multirow}
\usepackage[table,xcdraw]{xcolor} 
\usepackage{booktabs}
\usepackage{adjustbox}
\usepackage{pifont}
\usepackage{graphicx}
\usepackage{subcaption}
\usepackage{amssymb}

\title{Self-supervised Multiplex Consensus Mamba for General Image Fusion}
\author{
    Yingying Wang\textsuperscript{\rm 1},
    Rongjin Zhuang\textsuperscript{\rm 1},
    Hui Zheng\textsuperscript{\rm 1},
    Xuanhua He\textsuperscript{\rm 2},
    Ke Cao\textsuperscript{\rm 3},\\
    Xiaotong Tu\textsuperscript{\rm 1}\thanks{Corresponding Author.},
    Xinghao Ding\textsuperscript{\rm 1}
}
\affiliations{
     \textsuperscript{\rm 1} Key Laboratory of Multimedia Trusted Perception and Efficient Computing,\\ Ministry of Education of China, Xiamen University, China\\
     \textsuperscript{\rm 2}The Hong Kong University of Science and Technology\\
     \textsuperscript{\rm 3}University of Science and Technology of China\\
    
    {wangyingying7@stu.xmu.edu.cn, \{xttu, dxh\}@xmu.edu.cn}\\

}

\usepackage{bibentry}

\begin{document}

\maketitle

\begin{abstract}
Image fusion integrates complementary information from different modalities to generate high-quality fused images, thereby enhancing downstream tasks such as object detection and semantic segmentation. Unlike task-specific techniques that primarily focus on consolidating inter-modal information, general image fusion needs to address a wide range of tasks while improving performance without increasing complexity. To achieve this, we propose SMC-Mamba, a Self-supervised Multiplex Consensus Mamba framework for general image fusion. Specifically, the Modality-Agnostic Feature Enhancement (MAFE) module preserves fine details through adaptive gating and enhances global representations via spatial-channel and frequency-rotational scanning. The Multiplex Consensus Cross-modal Mamba (MCCM) module enables dynamic collaboration among experts, reaching a consensus to efficiently integrate complementary information from multiple modalities. The cross-modal scanning within MCCM further strengthens feature interactions across modalities, facilitating seamless integration of critical information from both sources. Additionally, we introduce a Bi-level Self-supervised Contrastive Learning Loss (BSCL), which preserves high-frequency information without increasing computational overhead while simultaneously boosting performance in downstream tasks. Extensive experiments demonstrate that our approach outperforms state-of-the-art (SOTA) image fusion algorithms in tasks such as infrared-visible, medical, multi-focus, and multi-exposure fusion, as well as downstream visual tasks.
\end{abstract}


\section{Introduction}
Due to hardware limitations, single sensors often fail to capture the full complexity of real-world scenes. Image fusion addresses this by integrating complementary information. This field can be categorized into multi-modal image fusion (MMIF), including infrared-visible (IVIF) and medical image (MDIF) fusion, and digital photographic image fusion (DPIF), which covers multi-focus (MFIF) and multi-exposure (MEIF) image fusion. 

In recent years, deep learning has become the dominant approach for image fusion~\cite{liu2024promptfusion, liu2024infrared, li2025umcfuse, zhang2025omnifuse}, mainly leveraging CNNs~\cite{wang2023learning} and Transformers~\cite{li2025a2rnet}. CNNs are effective at capturing local features but struggle with long-range dependencies due to limited receptive fields. Transformers address this with global self-attention, but suffer from high computational costs that scale quadratically with input size.
State Space Models (SSMs), particularly Mamba~\cite{gu2023mamba}, offer a compelling alternative. Mamba enables global context modeling with linear complexity, overcoming the limitations of both CNNs and Transformers. These strengths inspire us to explore Mamba for efficient and scalable image fusion.

Existing image fusion methods predominantly concentrate on single-task designs, limiting their generalization across diverse tasks. Each fusion task—IVIF, MDIF, MFIF, and MEIF—has distinct goals, yet all aim to preserve high-frequency textures and structural details. A dynamic architecture that adapts to varying modalities can better handle these differences. Mixture of Experts (MoE)~\cite{jordan1994moe} offers a promising solution by leveraging expert modules to address diverse objectives, improving fusion quality and supporting downstream vision tasks.

However, existing deep learning methods often emphasize low-frequency content, struggling to accurately capture fine-grained high-frequency details. This inherent bias~\cite{rahaman2019spectral, xu2020frequency} degrades visual quality and negatively impacts overall fusion performance. Moreover, the inefficiency of regularization strategies~\cite{xiao_FAFusion, fuoli2021fourier} may lead to the loss of critical high-frequency information, hindering the recovery of textures and edges in the results. To address these limitations, we propose SMC-Mamba, a Self-supervised Multiplex Consensus Mamba for general image fusion. This framework comprises three core designs: a Modality-Agnostic Feature Enhancement module (MAFE), a Multiplex Consensus Cross-modal Mamba module (MCCM), and the Bi-level Self-supervised Contrastive Learning Loss (BSCL). 

Initially, to achieve high-quality fusion results with abundant intricate details and boost performance in downstream tasks, we design the task-agnostic BSCL regularization loss, which reinforces high-frequency textures and structures without increasing complexity. Specifically, the high-frequency components of the fused images are drawn towards to those of the input modalities, while being pushed away from their low-frequency components at both the feature and pixel levels within the latent spaces.

To effectively handle diverse fusion tasks, we propose the MCCM module, which encourages diverse feature preferences and fusion strategies across experts, while enabling dynamically activated experts to collaborate and converge toward a unified representation, thereby providing reliable results for image fusion and downstream tasks. Additionally, unlike convolutions or self-attention, Mamba employs a scanning scheme to capture long-range dependencies in a content-aware manner. However, poorly designed scans may separate adjacent pixels in sequence, disrupting feature continuity. Existing methods focus mainly on spatial scanning~\cite{zhu2024vision} or single-modal scenarios~\cite{Fusionmamba1, Fusionmamba2}, neglecting spatial-channel interactions and cross-modal dependencies. To address this, we introduce a cross-modal scanning mechanism within each MCCM expert, enhancing inter-modal feature exchange and enabling seamless fusion of complementary cues.

Furthermore, although SSMs effectively capture long-range context, they often struggle with preserving local details. To address this, we introduce the MAFE module, which integrates local and global branches. The local branch uses a gating mechanism to adaptively extract fine-grained spatial features, while the global branch leverages Mamba with spatial-channel and frequency-rotational scanning to enhance global representations. This design captures long-range spatial-channel correlations and frequency relationships, enabling efficient modeling of global context while retaining local precision and enhancing unimodal feature representations.

In summary, the contributions of our work are as follows:

\begin{itemize}
    \item We propose SMC-Mamba, a Self-supervised Multiplex Consensus Mamba for general image fusion. This approach aims to dynamically and efficiently integrate complementary information from various modalities, flexibly handling different image fusion tasks.

    \item We devise the MCCM module, which promotes diverse feature preferences and fusion strategies across experts and enables activated experts to converge toward a unified representation, thereby providing reliable results for image fusion and downstream tasks.
        
    \item We design a novel self-supervised BSCL regularization loss that enhances the preservation of high-frequency information at both feature and pixel levels without increasing model complexity, while also improving performance in downstream visual tasks.
    
    \item We introduce the cross-modal scanning to exploit long-range cross-modal dependencies, strengthening feature interactions and facilitating the seamless integration of complementary and critical information from both modalities.

\end{itemize}

\begin{figure*}[ht]
\centering
\includegraphics[width=\textwidth]{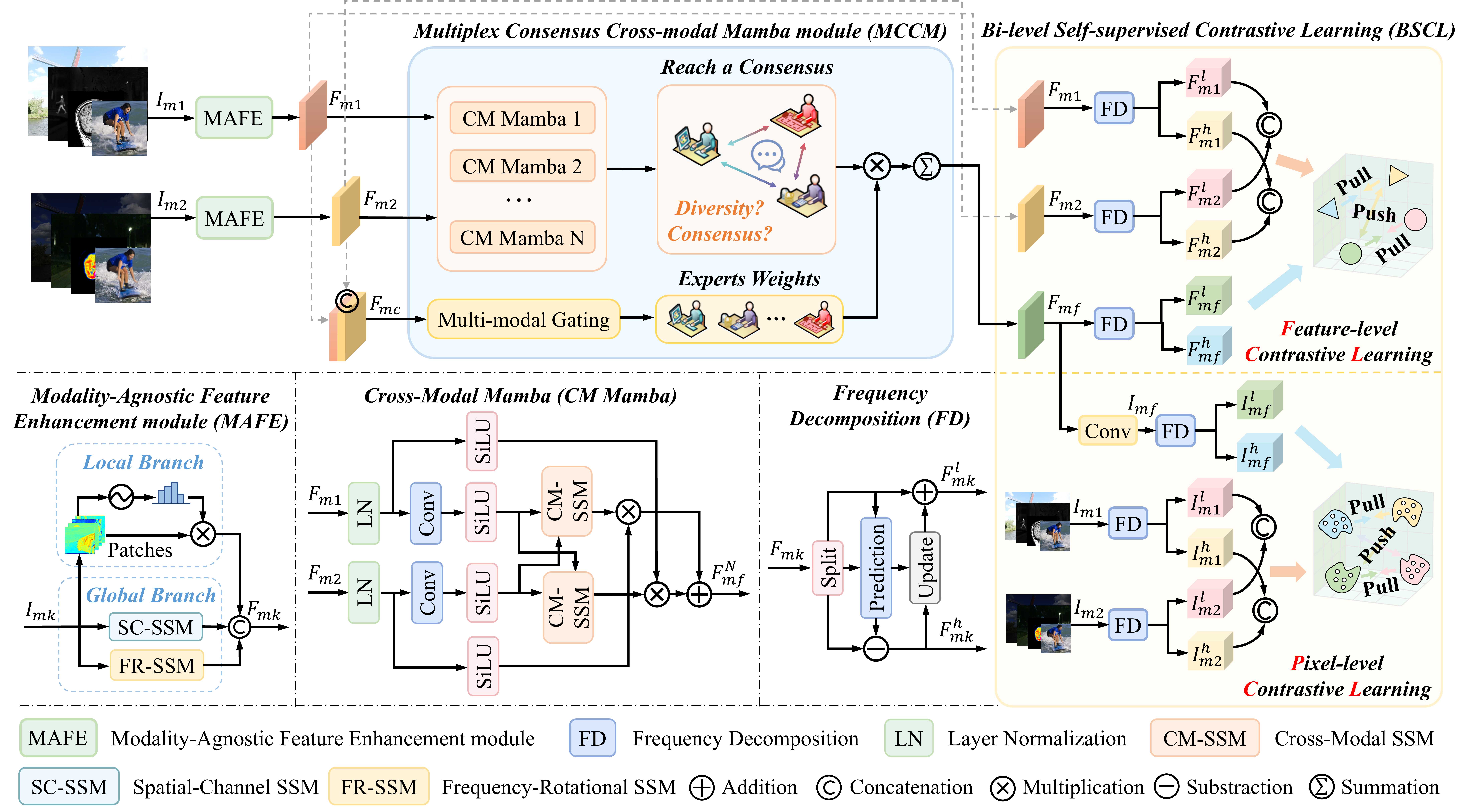}
\caption{\label{framework} The overall framework of our proposed network, which consists of three main components: 1) Modality-Agnostic Feature Enhancement module (MAFE). 2) Multiplex Consensus Cross-modal Mamba module (MCCM). 3) Bi-level Self-supervised Contrastive Learning Loss (BSCL).}
\end{figure*}

\section{Methodology}
In this section, we provide an in-depth overview of our proposed SMC-Mamba framework, as illustrated in Figure~\ref{framework}. The SMC-Mamba framework comprises three core components: MAFE, MCCM, and the BSCL approach. The details are illustrated as below.

\subsection{Modality-Agnostic Feature Enhancement}
Given source images $I_{mk} \in \mathbb{R}^{H \times W \times C_k}$ from tasks like IVIF, MDIF, MFIF, and MEIF (with modality index $k \in \{1, 2\}$), we extract shallow features $F_{sk}$ using a $3 \times 3$ convolution and layer normalization:
\begin{equation}
\begin{gathered}
F_{sk}=\text{LN}\left(\text{Conv}_{3 \times 3}\left(I_{mk}\right)\right).
\end{gathered}
\end{equation}

\textbf{Local Branch.} The shallow features $F_{sk} \in \mathbb{R}^{H \times W \times C}$ are first divided into patches $F_{sk}^j \in \mathbb{R}^{\frac{H}{2} \times \frac{W}{2} \times C}$ via tokenization. Each patch is processed with a $3 \times 3$ depth-wise convolution and then passed through a gating unit to adaptively capture local fine-grained details:
\begin{align}
    & F_{sk}^{j} = \text{Token}\left(F_{sk}\right), \\
    & F_{sk}^{j_{-}dw} = \text{DWConv}_{3 \times 3}\left(F_{sk}^{j}\right),
\end{align}
where $\text{Token}(\cdot)$ refers to the tokenization process, dividing the input shallow features $F_{sk}$ into smaller patches, and $j$ denotes the patch index.

Next, a GELU non-linearity~\cite{GELU} is applied to generate an attention map, which adaptively modulates $F_{sk}^{j_{-}dw}$ via element-wise multiplication:
\begin{equation}
\begin{gathered}
F_L = \operatorname{Gate}\left(\text{Conv}_{1 \times 1}\left(F_{sk}^{j_{-}dw}\right)\right) \odot F_{sk}^{j_{-}dw},
\end{gathered}
\end{equation}
where $\text{Conv}_{1 \times 1}(\cdot)$ denotes $1 \times 1$ convolution, $\operatorname{Gate}(\cdot)$ represents the gate function, and $\odot$ is the element-wise product.

\textbf{Global Branch.} In the spatial-channel SSM, input features $F_{sk}$ are fed into two parallel sub-branches: one applies a SiLU activation directly, while the other performs a $1 \times 1$ convolution followed by a $3 \times 3$ depth-wise convolution, both activated by SiLU. The outputs are then scanned using the spatial-channel scanning $\operatorname{SC\text{-}Scan}(\cdot)$:
\begin{align}
    & F_{DW} = \text{DWConv}_{3 \times 3}\left(\text{Conv}_{1 \times 1}\left(F_{sk}\right)\right),\\
    & F_{spa}^{sub1} = \text{LN}\left(\operatorname{SC{-}Scan}\left(\operatorname{SiLU}\left(F_{DW}\right)\right)\right), \\
    & F_{spa} = F_{spa}^{sub1} \odot \operatorname{SiLU}\left(F_{sk}\right).
\end{align}

In Fourier theory, modifying a single point in the frequency domain has a global impact on all input features. To enhance global representation, the frequency-rotational SSM processes $F_{sk}$ via two sub-branches: one applies SiLU activation directly, while the other transforms $F_{sk}$ into the frequency domain using the discrete Fourier transform (DFT):
\begin{equation}
\begin{gathered}
\mathcal{F}(F_{sk})(u, v)= \sum_{h=0}^{H-1} \sum_{w=0}^{W-1} F_{sk}(h, w) \cdot e^{-j 2 \pi\left(\frac{uh}{H}+\frac{vw}{W}\right)},
\end{gathered}
\end{equation}
where $u$ and $v$ denote the coordinates in the Fourier space, $\mathcal{F}(\cdot)$ represents the Fourier transformation.

The amplitude and phase components, $\mathcal{A}\left(F_{sk}\right)$ and $\mathcal{P}\left(F_{sk}\right)$, can be derived from the Fourier transform:
\begin{equation}
\begin{gathered}
\mathcal{A}\left(F_{sk}\right), \mathcal{P}\left(F_{sk}\right)=\mathcal{F}\left(F_{sk}\right).
\end{gathered}
\end{equation}

Then, a $3 \times 3$ depth-wise convolution and SiLU activation are applied to the amplitude and phase, followed by the frequency-rotational scanning $\operatorname{FR\text{-}Scan}(\cdot)$:
\begin{equation}
\begin{gathered}
F_{fre}^{\mathcal{A}} = \operatorname{FR{-}Scan}\left(\operatorname{SiLU}\left(\text{DWConv}_{3 \times 3}\left(\mathcal{A}\left(F_{sk}\right)\right)\right)\right),
\end{gathered}
\end{equation}
\begin{equation}
\begin{gathered}
F_{fre}^{\mathcal{P}} = \operatorname{FR{-}Scan}\left(\operatorname{SiLU}\left(\text{DWConv}_{3 \times 3}\left(\mathcal{P}\left(F_{sk}\right)\right)\right)\right).
\end{gathered}
\end{equation}

Next, the amplitude and phase features are transformed back to the spatial domain via inverse discrete Fourier transform (IDFT):
\begin{equation}
\begin{gathered}
F_{fre}=\mathcal{F}^{-1}\left(F_{fre}^{\mathcal{A}}, F_{fre}^{\mathcal{P}}\right) \odot \operatorname{SiLU}(F_{sk}),
\end{gathered}
\end{equation}
where $\mathcal{F}^{-1}(\cdot)$ denotes the IDFT operation.

After that, the global features can be derived as below:
\begin{equation}
\begin{gathered}
F_G=\operatorname{Cat}\left(F_{spa}, F_{fre}\right),
\end{gathered}
\end{equation}
where $\operatorname{Cat}(\cdot)$ is the concatenating function.

By integrating complementary local and global features, the MAFE module enhances modality-agnostic representation, enabling efficient long-range context capture while preserving local detail. The output features are as follows:
\begin{equation}
\begin{gathered}
F_{mk}=\operatorname{Cat}\left(F_{L}, F_{G}\right),
\end{gathered}
\end{equation}
where $k$ represents the index of each modality, with values of 1 and 2.

\textbf{Cross-modal Scanning.}
To enhance cross-modal feature interaction and aggregate complementary information, we propose cross-modal scanning $\operatorname{CM\text{-}Scan}(\cdot)$, comprising spatial and channel interaction scanning across modalities. Spatial scanning performs forward and reverse passes between modalities to model long-range spatial correlations, while channel scanning alternates across modalities to capture inter-modal dependencies. This strategy produce a more comprehensive and informative fused results. 

\begin{algorithm}[H]
\caption{Cross-modal Mamba Architecture\label{alg:cross-modal-mamba}}
\small
\textbf{Input:} Enhanced modality-agnostic features $F_{m1}$ and $F_{m2}$ \\
\textbf{Output:} Cross-modal Mamba fusion result $F_{mf}^N$

\begin{algorithmic}[1]
    \STATE \textcolor{gray}{\text{/* Layer normalization and reshape */}}
    \STATE $F_{ln1} \gets \operatorname{Linear}\left(\operatorname{LN}(F_{m1})\right)$
    \STATE $F_{ln2} \gets \operatorname{Linear}\left(\operatorname{LN}(F_{m2})\right)$
    \STATE \textcolor{gray}{\text{/* $1 \times 1$ convolution followed by SiLU activation */}}
    \STATE $F_{silu1} \gets \operatorname{SiLU}\left( \operatorname{Conv}_{1 \times 1}(F_{ln1}) \right)$
    \STATE $F_{silu2} \gets \operatorname{SiLU}\left( \operatorname{Conv}_{1 \times 1}(F_{ln2}) \right)$
    \STATE \textcolor{gray}{\text{/* Cross-modal scanning $\operatorname{CM-Scan}(\cdot)$ */}}
    \STATE $F_{cm1} \gets \operatorname{CM-Scan}(F_{silu1}, F_{silu2})$
    \STATE $F_{cm2} \gets \operatorname{CM-Scan}(F_{silu2}, F_{silu1})$
    \STATE \textcolor{gray}{\text{/* Cross-modal feature interactions and fusion */}}
    \STATE $F_{mf}^N \gets F_{cm1} \odot \operatorname{SiLU}(F_{ln2}) + F_{cm2} \odot \operatorname{SiLU}(F_{ln1})$
\end{algorithmic}
\textbf{Return} $F_{mf}^N$
\end{algorithm}

\begin{algorithm}[t]
\caption{Frequency Decomposition\label{alg:frequency-decomposition}}
\small
\textbf{Input:} Enhanced modality-agnostic features $F_{mk}$, fused feature $F_{mf}$, input images $I_{mk}$, and fused image $I_{mf}$ \\
\textbf{Output:} Feature-level low-frequency components $F_{mk}^l$ and $F_{mf}^l$, high-frequency residuals $F_{mk}^h$ and $F_{mf}^h$, image-level low-frequency components $I_{mk}^l$ and $I_{mf}^l$, high-frequency residuals $I_{mk}^h$ and $I_{mf}^h$

\begin{algorithmic}[1]
    \STATE \textcolor{gray}{\text{/* Feature-level. Channel-wise Split $\operatorname{S}(\cdot)$. */}}
    \STATE $F_{c1}, F_{c2} \gets \operatorname{S}(F_{mk})$ 
    \STATE $F_{cf1}, F_{cf2} \gets \operatorname{S}(F_{mf})$ 
    
    \STATE \textcolor{gray}{\text{/* Prediction  $\operatorname{P}(\cdot)$ for high-frequency residual */}}
    \STATE $F_{mk}^h \gets F_{c2} - \operatorname{P}(F_{c1})$
    \STATE $F_{mf}^h \gets F_{cf2} - \operatorname{P}(F_{cf1})$
    
    \STATE \textcolor{gray}{\text{/* Update $\operatorname{U}(\cdot)$ for low-frequency refinement */}}
    \STATE $F_{mk}^l \gets F_{c1} + \operatorname{U}(F_{mk}^h)$
    \STATE $F_{mf}^l \gets F_{cf1} + \operatorname{U}(F_{mf}^h)$
    
    \STATE \textcolor{gray}{\text{/* Image-level. Channel-wise Split $\operatorname{S}(\cdot)$. */}}
    \STATE $I_{c1}, I_{c2} \gets \operatorname{S}(I_{mk})$
    \STATE $I_{cf1}, I_{cf2} \gets \operatorname{S}(I_{mf})$
    
    \STATE \textcolor{gray}{\text{/* Prediction  $\operatorname{P}(\cdot)$ for high-frequency residual */}}
    \STATE $I_{mk}^h \gets I_{c2} - \operatorname{P}(I_{c1})$
    \STATE $I_{mf}^h \gets I_{cf2} - \operatorname{P}(I_{cf1})$
    
    \STATE \textcolor{gray}{\text{/* Update $\operatorname{U}(\cdot)$ for low-frequency refinement */}}
    \STATE $I_{mk}^l \gets I_{c1} + \operatorname{U}(I_{mk}^h)$
    \STATE $I_{mf}^l \gets I_{cf1} + \operatorname{U}(I_{mf}^h)$
\end{algorithmic}
\textbf{Return} $F_{mk}^h$, $F_{mk}^l$, $F_{mf}^h$, $F_{mf}^l$, $I_{mk}^h$, $I_{mk}^l$, $I_{mf}^h$, $I_{mf}^l$
\end{algorithm}

\subsection{Multiplex Consensus Cross-modal Mamba module}
To effectively capture complex cross-modal correlations, we propose the Multiplex Consensus Cross-modal Mamba (MCCM) module, which integrates multiple cross-modal Mamba experts $\{\operatorname{CM}_1, \dots, \operatorname{CM}_N\}$ under a unified gating framework. Each expert performs independent cross-modal fusion, while the gating network adaptively determines their importance based on input content.

Given modality-agnostic features $F_{mk}$ ($k \in \{1, 2\}$), we concatenate them into $F_{mc}$ and pass it through the gating network. Global Average Pooling (GAP) and Global Max Pooling (GMP) are first applied to extract representative global features:
\begin{align}
    F_{mc} &= \operatorname{Cat}(F_{m1}, F_{m2}), \\
    F_{g} &= \operatorname{GAP}(F_{mc}) + \operatorname{GMP}(F_{mc}).
\end{align}
A learnable noise term $\epsilon$ is added, controlled by $\operatorname{Softplus}(\cdot)$ to ensure non-negative noise for stable activation:
\begin{align}
    \epsilon &= \mathcal{N}(0,1) \cdot \operatorname{Softplus}(F_g \cdot W_{\text{noise}}).
\end{align}
The expert weights are computed as:
\begin{align}
    W_{\text{exp}} &= \operatorname{Softmax}\left(\operatorname{TopK}(F_g \cdot W_g + \epsilon)\right),
\end{align}
only the top-$k$ experts ($k=2$) are activated, the unselected experts receive zero weight. The added learnable noise introduces randomness, encouraging balanced expert selection.

During training, all experts are used with weights from $W_{\text{exp}}$ to guide learning. At inference, only the top-$k$ experts are executed, enabling efficient, task-adaptive computation.

Each expert follows a cross-modal Mamba architecture (Figure~\ref{framework}) that includes layer normalization, linear projection, a $1 \times 1$ convolution with SiLU activation, and the proposed cross-modal scanning operator $\operatorname{CM\text{-}Scan}(\cdot)$ to enable rich inter-modal interactions. The full process is detailed in Algorithm~\ref{alg:cross-modal-mamba}. The output of MCCM is the weighted sum of expert outputs:
\begin{align}
F_{mf} = \sum_{i=1}^{N} W_{\text{exp}}^i \cdot \operatorname{CM}_i(F_{mc}),
\end{align}
where $\operatorname{CM_{i}}(\cdot)$ represents the $i$-th cross-modal Mamba expert network. $N$ denotes the number of experts, with $N$ set to 4.

\textbf{Workload Balancing Loss.}
To prevent gating collapse and ensure all experts contribute during training, we introduce a load balancing loss based on the coefficient of variation:
\begin{align}
\mathcal{L}_{\text{wb}} = \left( \frac{\sigma(W_{\text{exp}})}{\overline{W_{\text{exp}}}} \right)^2,
\end{align}
where $\sigma(\cdot)$ and $\overline{(\cdot)}$ denote the standard deviation and mean of expert weights, respectively.

\textbf{Expert Diversity Loss.} To encourage heterogeneous expert behavior, we propose the expert diversity loss $\mathcal{L}_{\text{div}}$, which promotes diverse feature preferences and fusion strategies across expert, fostering a complementary and specialized ensemble:
\begin{align}
\mathcal{L}_{\text{div}} = \frac{1}{N(N - 1)} \sum_{i \ne j} \cos\left(\hat{F}_i, \hat{F}_j\right),
\end{align}
where $\hat{F}_i = \operatorname{CM}_i(F_{mc})$ is the output of the $i$-th cross-modal Mamba expert, $\cos(\hat{F}_i, \hat{F}_j)$ denotes the cosine similarity between expert outputs, $N$ is the total number of experts. Lower similarity indicates stronger diversity.

\textbf{Consensus Loss.}
To ensure consistent fusion outputs, we also encourage the activated experts to converge toward a unified representation, thereby providing reliable results for
image fusion and downstream tasks. The consensus feature is computed as the weighted average of expert outputs:
\begin{align}
F_{\text{consensus}} = \sum_{i=1}^{N} W_{\text{exp}}^i \cdot \hat{F}_i.
\end{align}
The consensus loss $\mathcal{L}_{\text{cons}}$ penalizes deviations from this aggregated representation:
\begin{align}
\mathcal{L}_{\text{cons}} = \sum_{i=1}^{N} W_{\text{exp}}^i \cdot \left\| \hat{F}_i - F_{\text{consensus}} \right\|_2^2.
\end{align}

\paragraph{Joint Objective.}
To balance expert specialization and collaboration, we combine these objectives with a time-decayed weighting scheme:
\begin{align}
\mathcal{L}_{\text{mccm}} = \mathcal{L}_{\text{wb}} + \lambda(t) \cdot \mathcal{L}_{\text{div}} + (1 - \lambda(t)) \cdot \mathcal{L}_{\text{cons}},
\end{align}
where $\lambda(t) = \cos\left(\frac{t}{T} \cdot \frac{\pi}{2}\right)$ decays over epochs ($t$ is the current epoch, $T$ denotes the total epochs), prioritizing diversity in the early stages and consensus in later stages. This dynamic balance enables the expert ensemble to first explore diverse fusion strategies and then consolidate into robust and aligned representations.

\begin{figure*}[t]
  \centering
  \includegraphics[width=0.9\linewidth]{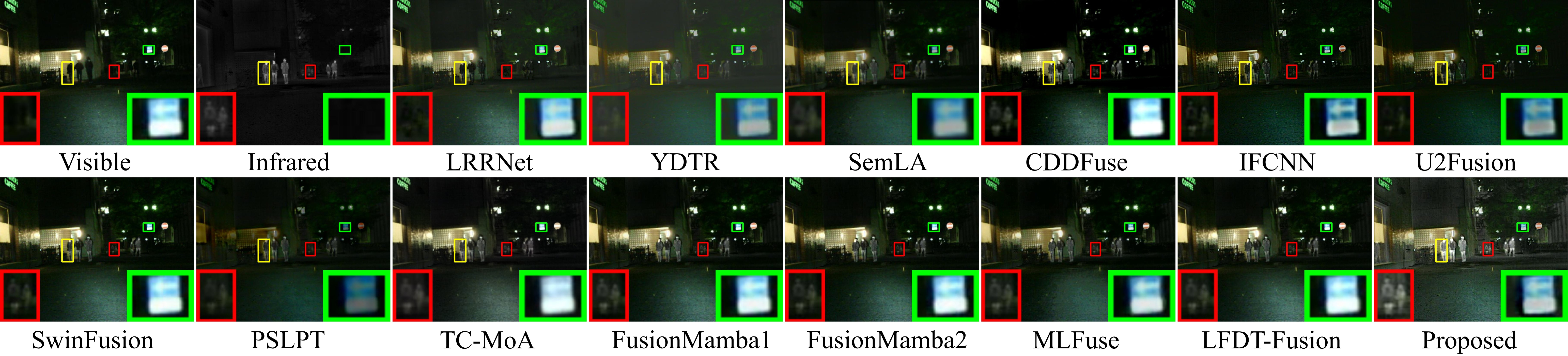}
  \caption{Visual comparisons of all the compared approaches on the MSRS dataset in IVIF task.}
  \label{fig:MSRS}
\end{figure*}

\begin{figure*}[t]
  \centering
  \includegraphics[width=0.9\linewidth]{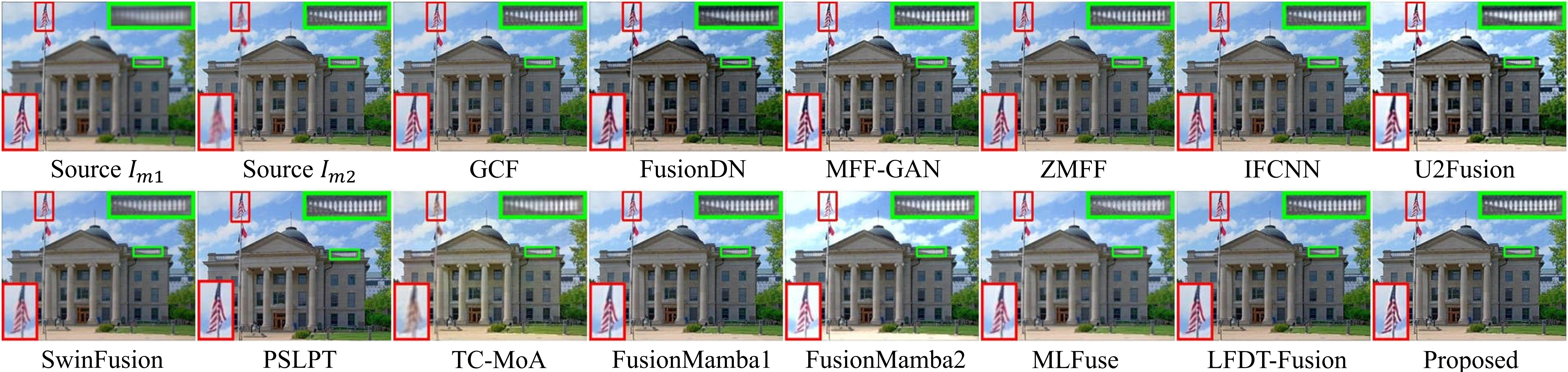}
  \caption{Visual comparisons of all the compared approaches on the MFI-WHU dataset in MFIF task.}
  \label{fig:MFI-WHU}
\end{figure*}

\subsection{Bi-level Self-supervised Contrastive Learning Loss}
For general image fusion, enhancing high-frequency detail without increasing model complexity remains challenging. To tackle this, we propose a Bi-level Self-supervised Contrastive Learning Loss (BSCL) that constrains high-frequency representations at both feature and pixel levels.

Specifically, we use the Haar wavelet lifting scheme~\cite{waveletslifting} to decompose fused and modality-enhanced features into high- and low-frequency components, as shown in Figure~\ref{framework}. The enhanced modality-agnostic feature $F_{mk}$ is split into two subsets, $F_{c1}$ and $F_{c2}$, via a channel-wise split operation $\operatorname{S}(\cdot)$.

Since $F_{c1}$ and $F_{c2}$ originate from the same source, they are strongly correlated. The Prediction block $\operatorname{P}(\cdot)$ uses the coarse low-frequency component $F_{c1}$ to predict the fine-grained high-frequency $F_{c2}$, yielding the high-frequency residual $F_{mk}^h$. The Update block $\operatorname{U}(\cdot)$ then refines $F_{c1}$ using feedback from $F_{mk}^h$, producing the updated low-frequency component $F_{mk}^l$.

A similar decomposition is applied to the fused feature $F_{mf}$, generating $F_{mf}^h$ and $F_{mf}^l$. At the image level, the fused image $I_{mf}$ and source images $I_{mk}$ are also decomposed using the Haar wavelet lifting scheme. The complete process is outlined in Algorithm~\ref{alg:frequency-decomposition}.

\textbf{Feature-level Contrastive Learning.}
Given the fused feature $F_{mf}$ and the enhanced modality-agnostic features $F_{mk}$, BSCL aims to pull the fused high-frequency components $F_{mf}^h$ closer to $F_{mk}^h$ while pushing them away from the low-frequency components $F_{mk}^l$ in latent space. We begin by concatenating the high- and low-frequency components of the input modalities:
\begin{align}
    & F_{mc}^h = \operatorname{Cat}\left(F_{m1}^h, F_{m2}^h\right), \\
    & F_{mc}^l = \operatorname{Cat}\left(F_{m1}^l, F_{m2}^l\right).
\end{align}

Then, the feature-level contrastive constraint is defined as:
\begin{equation}
\begin{gathered}
\mathcal {L}_{\text{fcl}}=\frac{\left\|F_{mf}^h-F_{mc}^h\right\|_1^2}{\left\|F_{mf}^h-F_{mc}^l\right\|_1^2}+\frac{\left\|F_{mf}^l-F_{mc}^l\right\|_1^2}{\left\|F_{mf}^l-F_{mc}^h\right\|_1^2}.
\end{gathered}
\end{equation}

\textbf{Pixel-level Contrastive Learning.}
Similarly, given the fused image $I_{mf}$ and input images $I_{mk}$, pixel-level contrastive learning pulls the fused high-frequency components $I_{mf}^h$ closer to $I_{mk}^h$ and pushes them away from $I_{mk}^l$. We first concatenate the high and low-frequency components of the input images:
\begin{align}
    & I_{mc}^h = \operatorname{Cat}\left(I_{m1}^h, I_{m2}^h\right), \\
    & I_{mc}^l = \operatorname{Cat}\left(I_{m1}^l, I_{m2}^l\right).
\end{align}

The pixel-level contrastive constraint is defined as:
\begin{equation}
\begin{gathered}
\mathcal {L}_{\text{pcl}}=\frac{\left\|I_{mf}^h-I_{mc}^h\right\|_1^2}{\left\|I_{mf}^h-I_{mc}^l\right\|_1^2}+\frac{\left\|I_{mf}^l-I_{mc}^l\right\|_1^2}{\left\|I_{mf}^l-I_{mc}^h\right\|_1^2}.
\end{gathered}
\end{equation}

\subsection{Overall Loss Function}
The overall loss function is defined as follows:
\begin{align}
\mathcal {L}_{\text{total}} &= \lambda_1 \mathcal {L}_{\text{fcl}} + \lambda_2 \mathcal {L}_{\text{pcl}} + \lambda_3 \mathcal{L}_{\text{mccm}} + \lambda_4 \mathcal {L}_{\text{ssim}} + \lambda_5 \mathcal {L}_{\text{int}},
\end{align}
where the hyperparameters $\lambda_1$ to $\lambda_5$ control the contribution of each sub-loss term and are empirically set to 0.8, 0.4, 1, 1, and 1, respectively.  
$\mathcal {L}_{\text{ssim}}$ denotes the SSIM loss~\cite{wang2004ssim}, and $\mathcal {L}_{\text{int}}$ represents the intensity loss as introduced in~\cite{zhang2020ifcnn}.

\section{Experiment}\label{sec:result}

\subsection{Implementation Details}
We implement our model using PyTorch and train it on a single NVIDIA RTX 3090 GPU. The ADAM optimizer with $\beta = 0.9$ is used with a batch size of 1 and an initial learning rate of $2 \times 10^{-4}$, which is halved every 1000 iterations via cosine annealing. In MCCM, we use $N = 4$ cross-modal Mamba experts.

\subsection{Datasets}
For the IVIF task, we train on the MSRS~\cite{MSRS_dataset} dataset and test on MSRS, RoadScene~\cite{FusionDN}, and M$^3$FD~\cite{M3FD_dataset}. MSRS and M$^3$FD are also used for downstream detection evaluation, while MSRS is used for segmentation.
For medical image fusion, we utilize the Harvard medical dataset, which includes CT-MRI, PET-MRI, and SPECT-MRI tasks, each used independently for both training and testing.
For multi-focus fusion, the MFI-WHU~\cite{MFF-GAN} dataset is used for training, with testing on both Lytro~\cite{Lytro} and MFI-WHU.
For multi-exposure fusion, we train on the MEF~\cite{MEF_dataset} dataset and test on the MEF benchmark~\cite{MEF_benchmark}.

\subsection{Comparison Methods and Evaluation Metrics}
We conduct comparisons with several SOTA techniques, including both general image fusion frameworks and task-specific approaches. Specifically, nine unified image fusion frameworks include IFCNN~\cite{zhang2020ifcnn}, U2Fusion~\cite{xu2020u2fusion}, SwinFusion~\cite{ma2022swinfusion}, PSLPT~\cite{wang2024general}, TC-MoA~\cite{zhu2024task}, Fusionmamba1~\cite{Fusionmamba1}, Fusionmamba2~\cite{Fusionmamba2}, MLFuse~\cite{MLFuse}, and LFDT-Fusion~\cite{LFDT-Fusion}. In addition, we also compare with task-specific methods. LRRNet~\cite{li2023lrrnet}, YDTR~\cite{Tang_2023_YDTR}, SemLA~\cite{xie2023semla}, and CDDFuse~\cite{zhao2023cddfuse} for IVIF task. EMFusion~\cite{xu2021emfusion}, MSRPAN~\cite{fu2021multiscale}, TUFusion~\cite{zhao2023transformer} and ALMFnet~\cite{mu2024almfnet} for MDIF. GCF~\cite{GCF}, FusionDN~\cite{FusionDN}, MFF-GAN~\cite{MFF-GAN} and ZMFF~\cite{ZMFF} for MFIF. DPE-MEF~\cite{han2022multi}, AGAL~\cite{liu2022agal}, BHF-MEF~\cite{mu2023little} and SAMT-MEF~\cite{huang2024leveraging} for MEIF task.


\begin{table}[!t]
    \centering
    \huge
    \resizebox{\linewidth}{!}{
    \begin{tabular}{lc|c|cccccccc}
    \toprule
    \midrule
        \multicolumn{3}{c}{\textbf{Methods}} & MI$\uparrow$ & SF$\uparrow$ & AG$\uparrow$ & CC$\uparrow$ & SCD$\uparrow$ & VIF$\uparrow$ &  $Q_{\mathrm{abf}}\uparrow$ & MS\_SSIM$\uparrow$ \\ 
         \midrule
         \multirow{14}{*}{\rotatebox{90}{MSRS}} & \multirow{4}{*}{\rotatebox{90}{Task-spec}} & LRRNet & 2.922 & 8.472 & 2.651 & 0.515 & 0.791 & 0.541 & 0.454 & 0.373\\ 
        ~ & ~ & YDTR & 2.760 & 7.404 & 2.201 & 0.631 & 1.138 & 0.577 & 0.349 & 0.441\\
        ~ & ~ & SemLA & 2.442 & 6.339 & 2.239 & \underline{0.641} & 1.392 & 0.608 & 0.290 & 0.498\\ 
        ~ & ~ & CDDFuse & 3.657 & 12.083 & \underline{4.043} & 0.596 & 1.549 & 0.819 & 0.548 & 0.459\\ 
        \cline{2-11}
        ~ & \multirow{10}{*}{\rotatebox{90}{General}} & IFCNN & 1.796 & \underline{12.134} & 4.030 & {0.633} & 1.374 & 0.579 & 0.479 & 0.504\\ 
        ~ & ~ & U2Fusion & 2.183 & 9.242 & 2.899 & 0.632 & 1.258 & 0.512 & 0.391 & 0.440\\ 
        ~ & ~ & SwinFusion & 3.652 & 11.038 & 3.546 & 0.595 & 1.647 & 0.825 & 0.558 & 0.504\\ 
        ~ & ~ & PSLPT & 2.284 & 10.419 & 3.306 & 0.610 & 1.374 & 0.753 & 0.553 & 0.501\\
        ~ & ~ & TC-MoA & 3.251 & 9.370 & 3.251 & 0.613 & \underline{1.661} & 0.811 & 0.565 & 0.515\\
        ~ & ~ & Fusionmamba1 & 4.121 & 10.955 & 3.599 & 0.611 & 1.635 & \underline{0.974} & \underline{0.652} & 0.511\\
        ~ & ~ & Fusionmamba2 & 3.608 & 11.401 & 3.658 & 0.610 & 1.645 & 0.947 & 0.637 & \underline{0.520}\\
        ~ & ~ & MLFuse & 2.889 & 8.819 & 2.962 & 0.634 & 1.520 & 0.753 & 0.519 & 0.498 \\
        ~ & ~ & LFDT-Fusion & \underline{4.216} & 11.236 & 3.694 & 0.600 & 1.637 & 0.876 & 0.624 & 0.512 \\
        ~ & ~ & \textbf{Proposed} & \textbf{4.490} & \textbf{12.211} & \textbf{4.054} & \textbf{0.699} & \textbf{1.664} & \textbf{0.991} & \textbf{0.658} & \textbf{0.522}\\ 
    \hline
    \multirow{14}{*}{\rotatebox{90}{RoadScene}} & \multirow{4}{*}{\rotatebox{90}{Task-spec}} & LRRNet & 2.704 & 11.114 & 4.166 & 0.621 & 1.430 & 0.488 & 0.323 & 0.537 \\ 
        ~ & ~ & YDTR & 3.043 & 10.788 & 4.035 & 0.591 & 1.229 & 0.602 & 0.463 & 0.524 \\ 
        ~ & ~ & SemLA & 2.808 & 15.571 & 4.899 & 0.606 & 1.269 & 0.564 & 0.415 & 0.518 \\ 
        ~ & ~ & CDDFuse & 3.001 & \textbf{19.779} & \textbf{7.029} & 0.623 & \underline{1.707} & 0.610 & 0.450 & 0.515  \\ 
        \cline{2-11}
        ~ & \multirow{10}{*}{\rotatebox{90}{General}} & IFCNN & 2.842 & 15.994 & 6.304 & 0.637 & 1.558 & 0.591 & 0.536 & 0.542 \\ 
        ~ & ~ & U2Fusion & 2.578 & 15.282 & 6.099 & 0.630 & 1.605 & 0.564 & 0.506 & \underline{0.546} \\
        ~ & ~ & SwinFusion & 3.334 & 12.161 & 4.516 & 0.623 & 1.576 & 0.614 & 0.450 & 0.534 \\ 
        ~ & ~ & PSLPT & 2.001 & 9.172 & 3.639 & 0.525 & 1.009 & 0.134 & 0.171 & 0.238\\
        ~ & ~ & TC-MoA & 2.853 & 12.786 & 5.339 & 0.611 & 1.562 & 0.577 & 0.477 & 0.522\\
        ~ & ~ & Fusionmamba1 & 3.189 & 14.659 & 5.602 & 0.632 & 1.322 & \underline{0.635} & \underline{0.543} & 0.519 \\
        ~ & ~ & Fusionmamba2 & 3.213 & 15.844 & 5.711 & 0.624 & 1.580 & 0.621 & 0.496 & 0.538\\
        ~ & ~ & MLFuse & 2.948 & 13.272 & 5.094 & \underline{0.640} & 1.595 & 0.629 & 0.527 & 0.545\\
        ~ & ~ & LFDT-Fusion & \underline{3.642} & 13.997 & 5.215 & 0.623 & 1.209 & 0.624 & 0.529 & 0.523\\
        ~ & ~ & \textbf{Proposed} & \textbf{3.772} & \underline{17.971} & \underline{6.866} & \textbf{0.643} & \textbf{1.733} & \textbf{0.642} & \textbf{0.557} & \textbf{0.547}\\ 
        \hline
        \multirow{14}{*}{\rotatebox{90}{M$^3$FD}} & \multirow{4}{*}{\rotatebox{90}{Task-spec}} & LRRNet & 2.892 & 11.162 & 3.700 & 0.522 & 1.726 & 0.556 & 0.510 & 0.418\\ 
        ~ & ~ & YDTR & 3.034 & 7.586 & 2.748 & 0.521 & 1.509 & 0.470 & 0.302 & 0.477\\ 
        ~ & ~ & SemLA & 2.376 & 7.285 & 3.181 & 0.480 & 1.495 & 0.542 & 0.363 & 0.473\\ 
        ~ & ~ & CDDFuse & 3.994 & \underline{17.578} & \underline{5.706} & 0.511 & 1.673 & 0.802 & 0.613 & 0.460\\ 
        \cline{2-11}
        ~ & \multirow{10}{*}{\rotatebox{90}{General}} & IFCNN & 2.630 & 16.250 & 5.448 & 0.554 & 1.710 & 0.685 & 0.590 & 0.445\\ 
        ~ & ~ & U2Fusion & 2.683 & 14.248 & 5.179 & 0.539 & \underline{1.753} & 0.673 & 0.578 & 0.463\\ 
        ~ & ~ & SwinFusion & 4.020 & 14.415 & 4.798 & 0.500 & 1.588 & 0.746 & 0.616 & 0.492\\ 
        ~ & ~ & PSLPT & \textbf{4.563} & 6.439 & 2.107 & 0.367 & 0.638 & \underline{0.958} & 0.321 & 0.483 \\
        ~ & ~ & TC-MoA & 2.856 & 11.221 & 4.010 & 0.506 & 1.556 & 0.579 & 0.508 & 0.466\\
        ~ & ~ & Fusionmamba1 & 4.044 & 14.042 & 4.689 & 0.465 & 1.414 & 0.747 & 0.580 & 0.480\\
        ~ & ~ & Fusionmamba2 & 3.823 & 14.933 & 4.913 & 0.492 & 1.540 & 0.744 & 0.600 & 0.496\\
        ~ & ~ & MLFuse & 2.897 & 10.229 & 3.382 & \underline{0.560} & 1.600 & 0.592 & 0.460 & \underline{0.501}\\
        ~ & ~ & LFDT-Fusion & 3.920 & 15.040 & 4.958 & 0.446 & 1.352 & 0.874 & \underline{0.624} & 0.486\\
        ~ & ~ & \textbf{Proposed} & \underline{4.280} & \textbf{19.495} & \textbf{6.378} & \textbf{0.561} & \textbf{1.791} & \textbf{0.972} & \textbf{0.632} & \textbf{0.507} \\
        \midrule
        \bottomrule
\end{tabular}}
\caption{Average metrics of all methods on the IVIF task. \textbf{Bold} and \underline{underlined} values indicate the best and second-best scores, respectively.}
\label{IVIF}
\end{table}

\begin{table}[!t]
    \centering
    \huge
    \resizebox{\linewidth}{!}{
    \begin{tabular}{lc|c|cccccccc}
    \toprule
    \midrule
        \multicolumn{3}{c}{\textbf{Methods}} & MI$\uparrow$ & SF$\uparrow$ & AG$\uparrow$ & CC$\uparrow$ & SCD$\uparrow$ & VIF$\uparrow$ & $N_{\mathrm{abf}}\downarrow$ & MS\_SSIM$\uparrow$ \\ 
         \midrule
         \multirow{14}{*}{\rotatebox{90}{Lytro}} & \multirow{4}{*}{\rotatebox{90}{Task-spec}} & GCF & \textbf{7.438} & 19.399 & 6.811 & 0.971 & 0.539 & 1.259 & 0.010 & 0.891  \\
         ~ & ~ & FusionDN & 5.793 & 17.129 & 6.359 & 0.917 & 0.511 & 1.007 & 0.030 & 0.866 \\
         ~ & ~ & MFF-GAN & 6.066 & \underline{21.037} & \underline{7.394} & 0.972 & 0.755 & 1.099 & 0.051 & 0.877 \\ 
         ~ & ~ & ZMFF & 6.630 & 18.770 & 6.715 & 0.971 & 0.442 & 1.175 & 0.028 & 0.890 \\ 
        \cline{2-11}
        ~ & \multirow{10}{*}{\rotatebox{90}{General}} & IFCNN & 6.896 & 19.398 & 7.254 & 0.967 & 0.606 & 1.258 & 0.026 & 0.835 \\ 
        ~ & ~ & U2Fusion & 5.787 & 19.634 & 6.840 & 0.973 & 0.546 & 1.255 & 0.060 & 0.890 \\ 
        ~ & ~ & SwinFusion & 6.149 & 16.941 & 6.116 & 0.873 & \textbf{0.837} & 1.069 & 0.027 & 0.862 \\ 
        ~ & ~ & PSLPT & 3.201 & 18.766 & 6.686 & 0.810 & 0.308 & 0.207 & 0.105 & 0.445 \\
        ~ & ~ & TC-MoA & 5.356 & 14.593 & 5.502 & 0.962 & 0.506 & 1.040 & 0.030 & 0.849\\
        ~ & ~ & Fusionmamba1 & 6.426 & 17.973 & 6.523 & 0.975 & 0.762 & 1.163 & 0.022 & 0.882\\
        ~ & ~ & Fusionmamba2 & 5.836 & 17.104 & 6.179 & 0.971 & 0.760 & 1.046 & 0.024 & 0.842\\
        ~ & ~ & MLFuse & 5.965 & 14.032 & 5.179 & \underline{0.981} & 0.684 & 1.028 & \underline{0.008} & 0.892\\
        ~ & ~ & LFDT-Fusion & 6.906 & 19.074 & 6.631 & 0.973 & 0.546 & \underline{1.264} & 0.016 & \underline{0.896}\\
        ~ & ~ & \textbf{Proposed} & \underline{7.081} & \textbf{23.785} & \textbf{8.191} & \textbf{0.989} & \underline{0.787} & \textbf{1.339} & \textbf{0.007} & \textbf{0.899}\\ 
    \hline
        \multirow{14}{*}{\rotatebox{90}{MFI-WHU}} & \multirow{4}{*}{\rotatebox{90}{Task-spec}} & GCF & \textbf{7.269} & 26.577 & 8.146 & 0.966 & 0.537 & \underline{1.326} & 0.073 & 0.942 \\ 
        ~ & ~ & FusionDN & 5.351 & 24.029 & 8.469 & 0.961 & 0.884 & 1.012 & 0.083 & 0.846 \\
        ~ & ~ & MFF-GAN & 5.684 & \underline{29.438} & \underline{9.447} & 0.961 & 0.964 & 1.120 & 0.089 & 0.900 \\ 
        ~ & ~ & ZMFF & 5.780 & 24.347 & 8.105 & 0.950 & 0.405 & 1.053 & 0.074 & 0.923 \\ 
        \cline{2-11}
        ~ & \multirow{10}{*}{\rotatebox{90}{General}} & IFCNN & 6.670 & 26.474 & 8.254 & 0.967 & 0.606 & 1.258 & 0.084 & 0.935 \\ 
        ~ & ~ & U2Fusion & 5.151 & 24.177 & 8.727 & 0.965 & \textbf{1.094} & 1.018 & 0.093 & 0.861 \\ 
        ~ & ~ & SwinFusion & 6.160 & 16.682 & 5.755 & \underline{0.979} & 0.418 & 1.123 & 0.111 & 0.932 \\ 
        ~ & ~ & PSLPT & 3.257 & 25.277 & 8.049 & 0.777 & 0.285 & 0.287 & 0.109 & 0.511 \\
        ~ & ~ & TC-MoA & 4.820 & 16.037 & 6.134 & 0.960 & 0.544 & 0.978 & \underline{0.072} & 0.891\\
        ~ & ~ & Fusionmamba1 & 5.854 & 22.311 & 7.653 & 0.974 & 0.957 & 1.125 & 0.076 & 0.922\\
        ~ & ~ & Fusionmamba2 & 5.371 & 23.218 & 7.536 & 0.966 & 0.964 & 1.024 & 0.081 & 0.848\\
        ~ & ~ & MLFuse & 5.581 & 20.500 & 6.686 & 0.977 & 0.801 & 1.044 & 0.080 & 0.924\\
        ~ & ~ & LFDT-Fusion & 6.649 & 25.316 & 8.041 & 0.971 & 0.597 & 1.270 & 0.073 & \underline{0.943}\\
        ~ & ~ & \textbf{Proposed} & \underline{6.890} & \textbf{35.669} & \textbf{10.929} & \textbf{0.985} & \underline{0.972} & \textbf{1.344} & \textbf{0.070} & \textbf{0.948} \\
        \midrule
        \bottomrule
    \end{tabular}}
    \caption{Average metrics of all methods on the MFIF task.}
    \label{MFIF}
\end{table}

\begin{table*}
    \centering
    \resizebox{\linewidth}{!}{%
    \huge
    \begin{tabular}{c|c|ccc|cccccccc}
    \toprule
    \midrule
        \multirow{2}*{\textbf{Ablation}} & \multirow{2}*{\textbf{Configuration}} & \multirow{2}*{\textbf{Params (M)}} & \multirow{2}*{\textbf{FLOPs (G)}} & \multirow{2}*{\textbf{Inference Time (ms)}} & \multicolumn{8}{c}{\textbf{MSRS Dataset}}\\ \cline{6-13}
        ~ & ~ & ~ & ~ & ~ & MI$\uparrow$ & SF$\uparrow$ & AG$\uparrow$ & CC$\uparrow$ & SCD$\uparrow$ & VIF$\uparrow$ & $Q_{\mathrm{abf}}\uparrow$ & MS-SSIM$\uparrow$ \\ \hline
        \textbf{Proposed} & - & 0.149 & 46.105 & 288.545 & \textbf{4.490} & 12.211 & 4.054 & \textbf{0.699} & \textbf{1.664} & \textbf{0.991} & \textbf{0.658} & \textbf{0.522} \\
        \hline
        \multirow{3}*{Core Operations}  & Mamba $\rightarrow$ Conv & 0.325 & 78.843 & 430.392 & 3.190 & 12.126 & 4.022 & 0.626 & 1.610 & 0.735 & 0.529 & 0.509 \\ 
        ~ & Mamba $\rightarrow$ Window Attention & 0.392 & 58.313 & 792.461 & 3.780 & 11.463 & 3.113 & 0.406 & 1.415 & 0.672 & 0.454 & 0.459 \\ 
        ~ & Mamba $\rightarrow$ Self Attention & 0.240 & 60.747 & 1271.691 & 3.710 & \textbf{12.387} & \textbf{4.180} & 0.601 & 1.630 & 0.834 & 0.588 & 0.518 \\ \hline
        \multirow{2}*{Main Modules} & MAFE Module $\rightarrow$ None & 0.041 & 14.260 & 226.355 & 2.384 & 12.073 & 4.023 & 0.638 & 1.544 & 0.803 & 0.548 & 0.515 \\ 
        ~ & MCCM Module $\rightarrow$ None & 0.125 & 38.606 & 164.867 & 2.202 & 10.048 & 3.426 & 0.544 & 1.392 & 0.702 & 0.496 & 0.453 \\ \hline
        \multirow{7}*{Loss Functions}  & w/o $\mathcal {L}_{\text{fcl}}$ & - & - & - & 3.914 & 11.147 & 3.717 & 0.585 & 1.546 & 0.946 & 0.624 & 0.517 \\
        ~ & w/o $\mathcal {L}_{\text{pcl}}$ & - & - & - & 3.870 & 10.952 & 3.627 & 0.572 & 1.522 & 0.937 & 0.613 & 0.511 \\
        ~ & w/o $\mathcal {L}_{\text{fcl}} \ \& \ \mathcal {L}_{\text{pcl}}$ & - & - & - & 3.721 & 10.823 & 3.580 & 0.565 & 1.482 & 0.925 & 0.601 & 0.503 \\
        ~ & w/o $\mathcal{L}_{\text{wb}}$ & - & - & - & 3.840 & 11.142 & 3.804 & 0.596 & 1.583 & 0.947 & 0.632 & 0.510 \\
        ~ & w/o $\mathcal{L}_{\text{div}}$ & - & - & - & 3.601 & 10.997 & 3.697 & 0.582 & 1.560 & 0.929 & 0.614 & 0.500 \\
        ~ & w/o $\mathcal{L}_{\text{cons}}$ & - & - & - & 3.702 & 11.060 & 3.727 & 0.590 & 1.571 & 0.938 & 0.626 & 0.506 \\
        ~ & w/o $\mathcal{L}_{\text{mccm}}$ & - & - & - & 3.466 & 10.891 & 3.643 & 0.563 & 1.504 & 0.906 & 0.598 & 0.496 \\    
        \hline
        \multirow{3}*{Scanning Schemes} & w/o Spatial-channel scanning & - & - & - & 4.106 & 11.381 & 3.587 & 0.618 & 1.554 & 0.936 & 0.641 & 0.516 \\ 
        ~ & w/o Frequency-rotational scanning & - & - & - & 4.350 & 11.942 & 4.021 & 0.620 & 1.515 & 0.963 & 0.642 & 0.513 \\ 
        ~ & w/o Cross-modal scanning & - & - & - & 3.965 & 11.191 & 3.538 & 0.557 & 1.470 & 0.896 & 0.601 & 0.504 \\  \hline
        \multirow{1}*{Scanning Directions} & Bi-direction $\rightarrow$ Single direction & - & - & - & 4.270 & 12.080 & 4.013 & 0.670 & 1.639 & 0.932 & 0.621 & 0.513 \\ 
        \midrule\bottomrule
    \end{tabular}%
    }
    \caption{Ablation study for SMC-Mamba on the MSRS dataset. ``A $\rightarrow$ B'' means replacing A with B. The thop library counts the number of parameters and FLOPs at a resolution of 480 $\times$ 640 pixels. Best results are highlighted in \textbf{bold}.}
    \label{Ablation}
\end{table*}

For evaluation metrics, we select several non-reference metrics to measure the fusion results, including mutual information (MI), spatial frequency (SF), average gradient (AG), correlation coefficient (CC), sum of the correlations of differences (SCD), visual information fidelity (VIF), edge based similarity measurement (\(Q_{\mathrm{abf}}\)), multi-scale structural similarity index measure (MS-SSIM), and noise or artifacts added in fused image due to fusion process (\(N_{\mathrm{abf}}\)). 

\begin{table}[!t]
    \centering
    \resizebox{\linewidth}{!}{%
    \LARGE
    \begin{tabular}{c|c|cccccccc}
    \toprule
        \multicolumn{2}{c}{\textbf{Methods}} & Background & Car & Person & Bike & Curve & Barrier & mIoU \\ 
    \midrule
        \multirow{2}{*}{Source} & IR & 97.9 & 85.0 & 51.0 & 69.7 & 51.3 & 68.9 & 70.6 \\ 
        ~ & VIS & 97.9 & 86.7 & 39.5 & 70.4 & 53.2 & 71.4 & 69.9 \\ 
    \cline{1-9}
        \multirow{4}{*}{Task-spec} & LRRNet & 98.3 & 88.9 & 67.7 & 69.1 & 51.9 & 71.5 & 74.6 \\ 
        ~ & YDTR & 98.5 & 89.6 & 72.0 & 70.9 & 62.0 & 73.3 & 77.7 \\
        ~ & SemLA & 98.4 & 89.6 & 70.8 & 70.0 & 58.2 & 75.0 & 77.0 \\ 
        ~ & CDDFuse & 98.5 & 89.7 & \textbf{74.2} & 71.4 & 63.8 & 73.7 & 78.6 \\ 
    \cline{1-9}
        \multirow{10}{*}{General} & IFCNN & 98.4 & 88.8 & 71.3 & 71.7 & 57.7 & 71.3 & 76.5 \\ 
        ~ & U2Fusion & 98.4 & 88.3 & 71.3 & 71.2 & 58.8 & 71.1 & 76.5 \\ 
        ~ & SwinFusion & \underline{98.6} & 89.9 & 73.6 & \underline{72.3} & \underline{64.7} & 73.3 & 78.7 \\ 
        ~ & PSLPT & {98.5} & 89.8 & 73.7 & 71.8 & 59.4 & \underline{75.7} & 78.2 \\
        ~ & TC-MoA & 98.5 & 89.8 & 72.6 & 70.8 & 63.8 & 74.3 & 78.3 \\
        ~ & Fusionmamba1 & 98.4 & 88.8 & 71.3 & 67.8 & 61.8 & 71.1 & 76.5\\
        ~ & Fusionmamba2 & 98.5 & 89.9 & 72.9 & 70.0 & 63.3 & 74.6 & 78.2\\
        ~ & MLFuse & 98.5 & 89.9 & 73.6 & 71.0 & 63.8 & \textbf{75.9} & 78.8\\
        ~ & LFDT-Fusion & 98.5 & \underline{89.9} & \underline{74.0} & 71.9 & 64.9 & 74.4 & \underline{78.9}\\
        ~ & \textbf{Proposed} & \textbf{98.7} & \textbf{90.0} & 73.7 & \textbf{72.6} & \textbf{65.6} & 75.0 & \textbf{79.3} \\
    \bottomrule
    \end{tabular}%
    }
    \caption{IoU(\%) values for DeepLabV3+ on MSRS dataset.}
    \label{segmentation}
\end{table}

\subsection{Quantitative Comparison with SOTA Methods}
Tables~\ref{IVIF} and ~\ref{MFIF} present the quantitative results for the IVIF and MFIF tasks. The IVIF task is evaluated on the MSRS, RoadScene, and M$^3$FD datasets, and the MFIF task is assessed on the Lytro and MFI-WHU datasets. 
Our proposed method consistently outperforms existing approaches across nearly all metrics and datasets.

\subsection{Visual Quality Comparison with SOTA Methods}
The visual comparisons for the IVIF task are provided in Figure~\ref{fig:MSRS}. Only our method clearly highlights pedestrian targets within the red box. Figure~\ref{fig:MFI-WHU} illustrates the MFIF fusion results. Our method preserves fine-grained textures, such as sharp railings and clear flag lines, while maintaining accurate color fidelity, demonstrating superior visual quality.

\begin{figure}[!t]
  \centering
  \includegraphics[width=\linewidth]{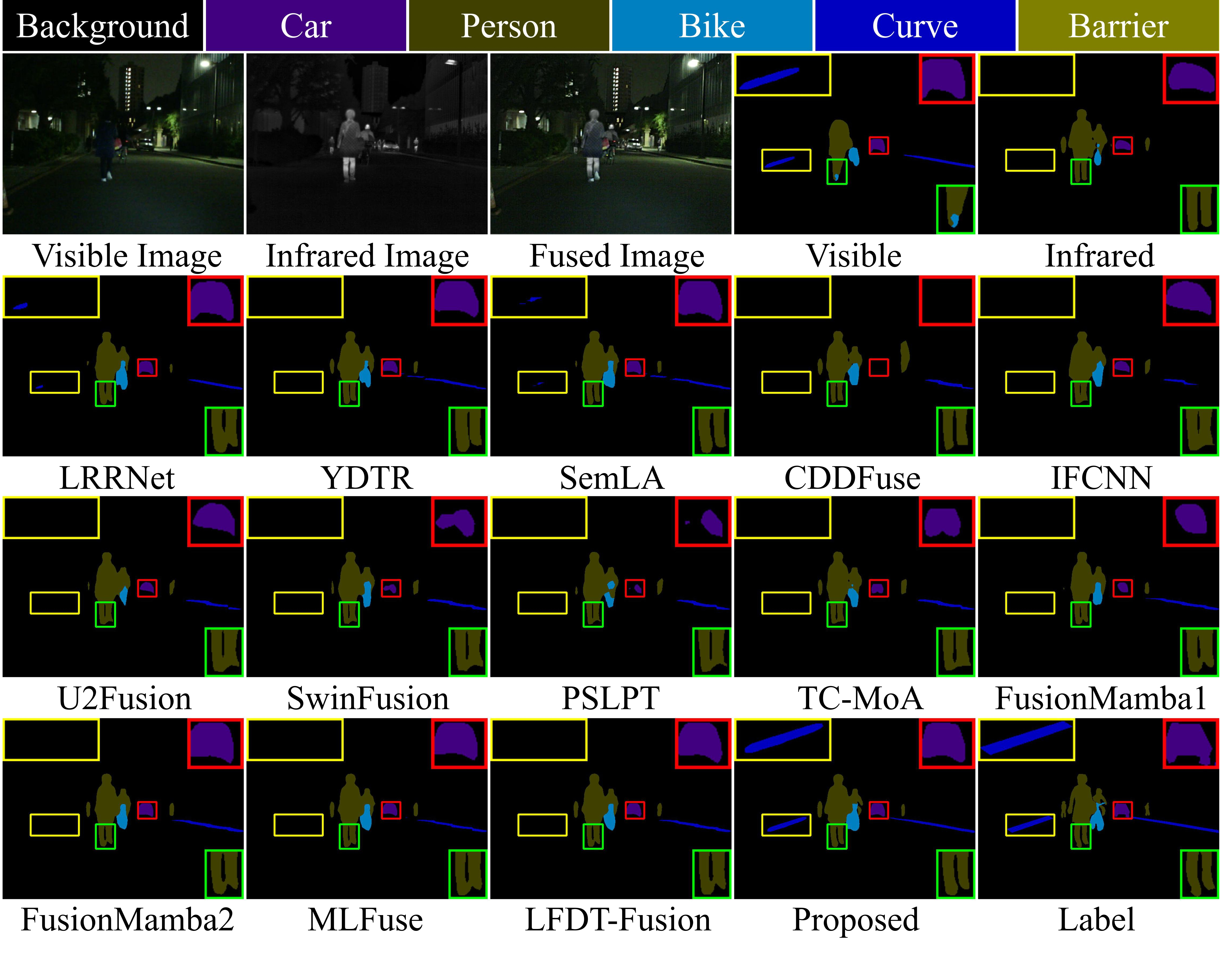}
  \caption{Qualitative segmentation on the MSRS dataset.}
  \label{fig:MSRS_Segment}
\end{figure}

\subsection{Ablation Study}
We conduct ablation studies on MSRS for the IVIF task to evaluate each core design, as shown in Table~\ref{Ablation}. The first part compares Mamba with commonly used operators: convolution layers, window attention, and self-attention. The second part assesses the proposed MAFE and MCCM modules by removing each one to evaluate its individual functionality. The third part evaluate the effectiveness of the feature-level contrastive loss $\mathcal {L}_{\text{fcl}}$, the pixel-level contrastive loss $\mathcal {L}_{\text{pcl}}$, the workload balancing loss $\mathcal {L}_{\text{wb}}$, the expert diversity loss $\mathcal{L}_{\text{div}}$, the consensus Loss $\mathcal{L}_{\text{cons}}$, and the MCCM loss $\mathcal{L}_{\text{mccm}}$. The fourth part validates the effectiveness of the scanning schemes, including spatial-channel scanning, frequency-rotational scanning, and cross-modal scanning. The fifth part examines the scanning directions, comparing single-directional scanning with bidirectional scanning.

\subsection{Downstream Tasks}
To investigate the benefits for downstream visual tasks, we present semantic segmentation results in Table~\ref{segmentation}. We employ the DeepLabV3+~\cite{DeepLabV3plus} to evaluate performance on the MSRS dataset. Our method achieves the highest mIoU value, demonstrating superior pixel-level segmentation accuracy. As shown in Figure~\ref{fig:MSRS_Segment}, our method produces the most accurate foot and car shapes and is the only one to correctly segment the roadside area.

\section{Conclusions}\label{sec:con}
In this paper, we introduce SMC-Mamba, a Self-supervised Multiplex Consensus Mamba for general image fusion. The MCCM module promotes diverse feature preferences and fusion strategies across experts and enables activated experts to converge toward a unified representation, thereby providing reliable results for image fusion and downstream tasks. The BSCL enhances the preservation of high-frequency details at both feature and pixel levels in a self-supervised manner. The cross-modal scanning captures cross-modal long-range dependencies, enabling seamless integration of complementary information. Meanwhile, MAFE boosts modality-agnostic features by capturing global context and preserving fine-grained local details. Qualitative and quantitative comparisons with the SOTA methods demonstrate the superiority of our proposed SMC-Mamba method. 

\section*{Acknowledgments} 
This work was supported by the National Natural Science Foundation of China under Grant 82272071, 62271430, 82172073, and 52105126.

\bibliography{main}

\end{document}